\documentclass{article}

    \usepackage[nonatbib,final]{neurips_2021}

\usepackage[utf8]{inputenc} 
\usepackage[T1]{fontenc}    
\usepackage{hyperref}       
\usepackage{url}            
\usepackage{booktabs}       
\usepackage{amsfonts}       
\usepackage{nicefrac}       
\usepackage{microtype}      
\usepackage{xcolor}         

\usepackage{amsmath}
\usepackage{amsfonts}
\usepackage{amsthm}
\usepackage{amssymb}
\usepackage{pdfpages}
\usepackage{subfigure}
\usepackage{booktabs}
\usepackage{float}
\usepackage{algorithm}
\usepackage{algorithmic}
\usepackage{adjustbox}
\usepackage{graphicx}
\usepackage{xcolor}

\newtheorem{theorem}{Theorem}[section]
\newtheorem{proposition}[theorem]{Proposition}
\newtheorem{lemma}[theorem]{Lemma}
\theoremstyle{definition}

\usepackage{color, soul}
\setulcolor{red} 
\setstcolor{blue} 
\sethlcolor{yellow} 
\numberwithin{equation}{section}

\hypersetup{
	colorlinks=true,
        breaklinks=true,
        citecolor={green!30!blue},
}

\title{MBDP: A Model-based Approach to Achieve both Robustness and Sample Efficiency via Double\\ Dropout Planning}

\author{%
  Wanpeng Zhang \\
  Tsinghua University\\
   \And
   Xi Xiao \\
   Tsinghua University \\
   \AND
   Yao Yao \\
   Tsinghua University \\
   \And
   Mingzhe Chen \\
   Tsinghua University \\
   \And
   Dijun Luo \\
   Tencent AI Lab \\
}

\begin{document}

\maketitle

\begin{abstract}
Model-based reinforcement learning is a widely accepted solution for solving excessive sample demands. However, the predictions of the dynamics models are often not accurate enough, and the resulting bias may incur catastrophic decisions due to insufficient robustness. Therefore, it is highly desired to investigate how to improve the robustness of model-based RL algorithms while maintaining high sampling efficiency. In this paper, we propose Model-Based Double-dropout Planning (MBDP) to balance robustness and efficiency. MBDP consists of two kinds of dropout mechanisms, where the rollout-dropout aims to improve the robustness with a small cost of sample efficiency, while the model-dropout is designed to compensate for the lost efficiency at a slight expense of robustness. By combining them in a complementary way, MBDP provides a flexible control mechanism to meet different demands of robustness and efficiency by tuning two corresponding dropout ratios. The effectiveness of MBDP is demonstrated both theoretically and experimentally.
\end{abstract}

\section{Introduction}
Reinforcement learning (RL) algorithms are commonly divided into two categories: model-free RL and model-based RL. Model-free RL methods learn a policy directly from samples {collected} in the real environment, while model-based RL approaches build approximate predictive models of the environment
to assist in the optimization of the policy \cite{chen2015reinforcement,polydoros2017survey}. In recent years, RL has achieved remarkable results in a wide range of areas, including continuous control \cite{schulman2015high,lillicrap2015continuous,levine2016end}, and outperforming human performances on Go and games \cite{mnih2015human,silver2016mastering}. However, most of these results are achieved by model-free RL algorithms, which rely on a large number of environmental samples for training, limiting the application scenarios when deployed in practice. In contrast, model-based RL methods have shown the promising potential to cope with the lack of samples by using predictive models for simulation and planning \cite{deisenroth2013survey,berkenkamp2017safe}. To reduce sample complexity, PILCO \cite{deisenroth2011pilco} learns a probabilistic model through Gaussian process regression, which models prediction uncertainty to boost agent's performance in complex environments. Based on PILCO, the DeepPILCO algorithm \cite{gal2016improving} enables the modeling of more complex environments by introducing the Bayesian Neural Network (BNN), a universal function approximator with high capacity. To further enhance the interpretability of the predictive models and improve the robustness of the learned policies \cite{Chua2018DeepModels,malik2019calibrated}, ensemble-based methods \cite{rajeswaran2016epopt,kurutach2018model} train an ensemble of models  to comprehensively capture the uncertainty in the environment and have been empirically shown to obtain significant improvements in sample efficiency \cite{levine2016end,Chua2018DeepModels,janner2019trust}.    

Despite the high sample efficiency, model-based RL methods inherently suffer from inaccurate predictions, especially when faced with high-dimensional tasks and insufficient training samples \cite{abbeel2006using,moerland2020model}. Model accuracy can greatly affect the policy quality, and policies learned in inaccurate models tend to have significant performance degradation due to cumulative model error \cite{sutton1996model,asadi2019combating}. Therefore, how to eliminate the effects caused by model bias has become a hot topic in model-based RL methods. Another important factor that limits the application of model-based algorithms is safety concerns. In a general RL setup, the agent needs to collect observations to extrapolate the current state before making decisions, which poses a challenge to the robustness of the learned policy because the process of acquiring observations through sensors may introduce random noise and the real environment is normally partial observable. Non-robust policies may generate disastrous decisions when faced with a noisy environment, and this safety issue is more prominent in model-based RL because the error in inferring the current state from observations may be further amplified by model bias when doing simulation and planning with the predictive models. Drawing on researches in robust control  \cite{zhou1998essentials}, a branch of control theory, robust RL methods have attracted more and more attention to improve the capability of the agent against perturbed states and model bias. The main objective of robust RL is to optimize the agent's performance in worst-case scenarios and to improve the generalization of learned policies to noisy environments \cite{shapiro2014lectures}. Existing robust RL methods can be roughly classified into two types, one is based on adversarial ideas such as RARL \cite{pinto2017robust} and NR-MDP \cite{tessler2019action} to obtain robust policies by proposing corresponding minimax objective functions, while the other group of approaches \cite{Tamar2015OptimizingSampling} introduce conditional value at risk (CVaR) objectives to ensure the robustness of the learned policies. However, the increased robustness of these methods can lead to a substantial loss of sample efficiency due to the pessimistic manner of data use. Therefore, it is nontrival to enhance the robustness of policy  while avoiding sample inefficiency.

In this paper, we propose \textbf{M}odel-\textbf{B}ased Reinforcement Learning with Double \textbf{D}ropout \textbf{P}lanning (MBDP) algorithm for the purpose of learning policies that can reach a balance between robustness and sample efficiency. Inspried by CVaR, we design the rollout-dropout mechanism to enhance robustness by optimizing the policies with low-reward samples. On the other hand, in order to maintain high sample efficiency and reduce the impact of model bias, we learn an ensemble of models to compensate for the inaccuracy of single model. Furthermore, when generating imaginary samples to assist in the optimization of policies, we design the model-dropout mechanism to avoid the perturbation of inaccurate models by only using models with small errors. To meet different demands of robustness and sample efficiency, a flexible control can be realized via the two dropout mechanisms. We demonstrate the effectiveness of MBDP both theoretically and empirically.

\section{Notations and Preliminaries}

\subsection{Reinforcement Learning}

We consider a Markov decision process (MDP), defined by the tuple $(\mathcal{S},\mathcal{A}, \mathcal{P}, r, \gamma)$, where $\mathcal{S}\in\mathbb{R}^{d_s}$ is the state space, $\mathcal{A}\in\mathbb{R}^{d_a}$ is the action space, $r(s,a):\mathcal{S}\times\mathcal{A}\mapsto\mathbb{R}$ is the reward function, $\gamma\in [0,1]$ is the discount factor, and $\mathcal{P}(s'|s,a):\mathcal{S}\times\mathcal{A}\times\mathcal{S}\mapsto[0,1]$ is the conditional probability distribution of the next state given current state $s$ and action $a$. The form $s'=\mathcal{P}(s,a) : \mathcal{S}\times\mathcal{A}\mapsto\mathcal{S}$ denotes the state transition function when the environment is deterministic. Let ${V}^{\pi,\mathcal{P}}(s)$ denote the expected return or expectation of accumulated rewards starting from initial state $s$, i.e., the expected sum of discounted rewards following policy $\pi(a|s)$ and state transition function $\mathcal{P}(s,a)$:
\begin{equation}\label{def:eta-s}
    {V}^{\pi,\mathcal{P}}(s) = \underset{\{a_0,s_1,\ldots\} \sim \pi,\mathcal{P}}{\mathbb{E}}\left[\sum_{t=0}^\infty\gamma^t r(s_t,a_t)\mid s_0=s\right]
\end{equation}

For simplicity of symbol, let ${V}^{\pi,\mathcal{P}}$ denote the expected return over random initial states:
\begin{equation}\label{def:eta-expectation}
    {V}^{\pi,\mathcal{P}} = \underset{s_0\in\mathcal{S}}{\mathbb{E}} \left[{V}^{\pi,\mathcal{P}}(s_0)\right]
\end{equation}

The goal of reinforcement learning is to maximize the expected return by finding the optimal decision policy, i.e., $\pi^* = {\arg\max}_\pi\ {V}^{\pi,\mathcal{P}}$.

\subsection{Model-based Methods}\label{sec:mb-methods}

In model-based reinforcement learning, an approximated transition model $\mathcal{M}(s,a)$ is learned by interacting with the environment, the policy $\pi(a|s)$ is then optimized with samples from the environment and data generated by the model. We use the parametric notation $\mathcal{M}_\phi, \phi\in\Phi$ to specifically denote the model trained by a neural network, where $\Phi$ is the parameter space of models.

More specifically, to improve the ability of models to represent complex environment, we need to learn multiple models and make an ensemble of them, i.e., $\mathcal{M}=\{\mathcal{M}_{\phi_1},\mathcal{M}_{\phi_2},\ldots\}$. To generate a prediction from the model ensemble, we select a model $\mathcal{M}_{\phi_i}$ from $\mathcal{M}$ uniformly at random, and perform a model rollout using the selected model at each time step, i.e., $s_{t+1}\sim \mathcal{M}_{\phi_t}(s_t,a_t)$. Then we fill these rollout samples $x=\left(s_{t+1},s_t,a_t\right)$ into a batch. Finally we can perform policy optimization on these generated samples.

\subsection{Conditional Value-at-Risk}\label{sec:CVaR-def}

Let $Z$ denote a random variable with a cumulative distribution function (CDF) $F(z)=\mathrm{Pr}(Z<z)$. Given a confidence level $p \in (0,1)$, the Value-at-Risk of $Z$ (at confidence level $p$) is denoted $\mathrm{VaR}_p(Z)$, and given by

\begin{equation}
    \mathrm{VaR}_p(Z)=F^{-1}(p)\triangleq \inf\{z:F(z)\geq p\}
\end{equation}

The Conditional-Value-at-Risk of $Z$ (at confidence level $p$) is denoted by $\mathrm{CVaR}_p(Z)$ and defined as the expected value of $Z$, conditioned on the $p$-portion of the tail distribution:

\begin{equation}\label{eq:def-cvar}
    \mathrm{CVaR}_p(Z)\triangleq \mathbb{E}[Z|Z\geq \mathrm{VaR}_p(Z)]
\end{equation}

\section{MBDP Framework}\label{sec:framework}

\begin{figure*}
\centering
\includegraphics[width=0.9\textwidth]{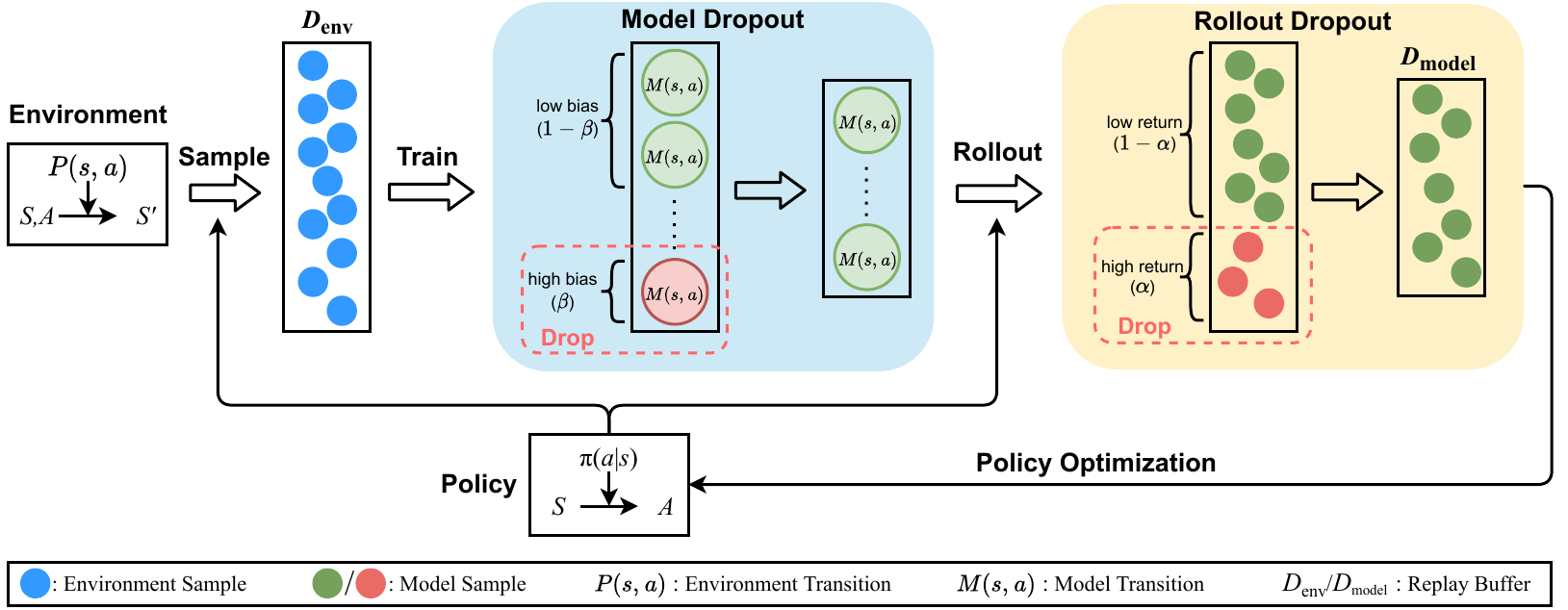}
\caption{Overview of the MBDP algorithm. When interacting with the environment, we collect samples into environment replay buffer $\mathcal{D}_{\mathrm{env}}$, used for training the simulator model of the environment. Then we implement the model-dropout procedure and perform rollouts on the model ensemble. The sampled data from the model ensemble is filled into a temporary batch, and then we get a dropout buffer $\mathcal{D}_\mathrm{model}$ by implementing the rollout-dropout procedure. Finally, we use samples from $\mathcal{D}_\mathrm{model}$ to optimize the policy $\pi(a|s)$.}
\label{fig:algo-structure}
\end{figure*}

In this section, we introduce how MBDP leverages Double Dropout Planning to find the balance between efficiency and robustness. The basic procedure of MBDP is to 1) sample data from the environment; 2) train an ensemble of models from the sampled data; 3) calculate model bias over observed environment samples, and choose a subset of model ensemble based on the calculated model bias; 4) collect rollout trajectories from the model ensemble, and make gradient updates based on the subsets of sampled data. The overview of the algorithm architecture is shown in figure \ref{fig:algo-structure} and the overall algorithm pseudo-code is demonstrated in Algorithm \ref{algo:our-method}.

We will also theoretically analyze robustness and performance under the dropout planning of our MBDP algorithm. For simplicity of theoretical analysis, we only consider deterministic environment and models in this section, but the experimental part does not require this assumption. The detailed proofs can be found in the appendix as provided in supplementary materials.

\begin{algorithm}[tb]
\caption{Model-Based Reinforcement Learning with Double Dropout Planning (\textbf{MBDP})}
\label{algo:our-method}
\begin{algorithmic}
\STATE Initialize hyperparameters, policy $\pi_\theta$, environment replay buffer $\mathcal{D}_{\mathrm{env}}$, model replay buffer $\mathcal{D}_{\mathrm{model}}$\\
\FOR{$N_\mathrm{epoch}$ iterations}
    \STATE Take an action in environment using policy $\pi_\theta$
    \STATE Add samples to $\mathcal{D}_{\mathrm{env}}$\\
    \FOR{$N_\mathrm{train}$ iterations}
        \STATE Train probabilistic model $\mathcal{M}$ on $\mathcal{D}_{\mathrm{env}}$\\
        \STATE Build a model subset $\mathcal{M}^\beta = \{\mathcal{M}_{\phi_1},\ldots,\mathcal{M}_{\phi_{N_{1-\beta}}}\}$ according to $\mathrm{bias}({\phi_i})$\\
        \FOR{$t=1,2,\ldots ,T$}
            \STATE Select a model $\mathcal{M}_{\phi_t}$ from $\mathcal{M}^\beta$ randomly\\
            \STATE Perform rollouts on model $\mathcal{M}_{\phi_t}$ with policy $\pi_\theta$ and get samples $x=\left(s_{t+1},s_t,a_t\right)$ \\
            \STATE Fill these samples into temp batch $\mathcal{B}^{\pi,\mathcal{M}^\beta}$\\
        \ENDFOR
        \STATE Calculate $r_{1-\alpha}(\mathcal{B}^{\pi,\mathcal{M}^\beta}|s)$: the $(1-\alpha)$ percentile of batch $\mathcal{B}^{\pi,\mathcal{M}^\beta}$ grouped by state $s, \forall s\in\mathcal{S}$\\
        \FOR{$x\in \mathcal{B}^{\pi,\mathcal{M}^\beta}$}
            \IF{$r(x)\leq r_{1-\alpha}(\mathcal{B}^{\pi,\mathcal{M}^\beta}|s_t)$}
                \STATE fill $x$ into $\mathcal{D}_{\mathrm{model}}$
            \ENDIF
        \ENDFOR
    \ENDFOR
    \STATE Optimize $\pi_\theta$ on $\mathcal{D}_{\mathrm{model}}$: $\theta\leftarrow \theta - \lambda\nabla_\theta J_\theta(\mathcal{D}_{\mathrm{model}})$
\ENDFOR
\end{algorithmic}
\end{algorithm}

\subsection{Rollout Dropout in MBDP}
\label{sec:dropout-remark}

Optimizing the expected return in a general way as model-based methods allows us to learn a policy that performs best in expectation over the training model ensemble. However, best expectation does not mean that the result policies can perform well at all times. This instability typically leads to risky decisions when facing poorly-informed states at deployment.

Inspired by previous works \cite{rajeswaran2016epopt,Tamar2015OptimizingSampling,chow2015risk} which optimize conditional value at risk (CVaR) to explicitly seek a robust policy, we add a dropout mechanism in the rollout procedure. Recall the model-based methods in Section \ref{sec:mb-methods}, to generate a prediction from the model ensemble, we select a model $\mathcal{M}_{\phi_i}$ from $\mathcal{M}$ uniformly at random, and perform a model rollout using the selected model at each time step, i.e., $s_{t+1}\sim \mathcal{M}_{\phi_t}(s_t,a_t)$. Then we fill these rollout samples $x=\left(s_{t+1},s_t,a_t\right)$ into a batch and retain a $(1-\alpha)$ percentile subset with more pessimistic rewards. We use $\mathcal{B}_\alpha^{\pi,\mathcal{M}}$ to denote the $(1-\alpha)$ percentile rollout batch:
\begin{equation}\label{def:batch-alpha-beta}
    \mathcal{B}_\alpha^{\pi,\mathcal{M}}=\left\{x|x\in\mathcal{B}^{\pi,\mathcal{M}},r(x|s)\leq r_{1-\alpha}(\mathcal{B}^{\pi,\mathcal{M}}|s), \forall s \in \mathcal{S}\right\}
\end{equation}
where $\mathcal{B}^{\pi,\mathcal{M}}=\left\{x|x\triangleq\left(s_{t+1},s_t,a_t\right)\sim\pi,\mathcal{M}\right\}$ and $r_{1-\alpha}(\mathcal{B}^{\pi,\mathcal{M}}|s)$ is the $(1-\alpha)$ percentile of reward values conditioned on state $s\in\mathcal{S}$ in batch $\mathcal{B}^{\pi,\mathcal{M}}$. The expected return of dropout batch rollouts is denoted by ${V}^{\pi,\mathcal{M}}_\alpha$:
\begin{equation}\label{def:eta-beta}
    {V}^{\pi,\mathcal{M}}_\alpha=\mathbb{E}\left[{\sum}_{\{s_0,a_0,\ldots\} \sim\mathcal{B}_\alpha^{\pi,\mathcal{M}}}\left[\gamma^t r(s_t,a_t)\right]\right]
\end{equation}

Rollout-dropout can improve the robustness with a nano cost of sample efficiency, we will analyze how it brings improvements to robustness in Section \ref{sec:guarantee}.

\subsection{Model Dropout in MBDP}

Rollout-dropout can improve the robustness, but it is clear that dropping a certain number of samples could affect the algorithm's sample efficiency. Model-based methods can improve this problem. However, since model bias can affect the performance of the algorithm, we also need to consider how to optimize it. Previous works use an ensemble of bootstrapped probabilistic transition models as in PETS method \cite{Chua2018DeepModels} to properly incorporate two kinds of uncertainty into the transition model. 


In order to mitigate the impact of discrepancies and flexibly control the accuracy of model ensemble, we design a model-dropout mechanism. More specifically, we first learn an ensemble of transition models $\{\mathcal{M}_{\phi_1},\mathcal{M}_{\phi_2},\ldots\}$, each member of the ensemble is a probabilistic neural network whose outputs $\mu_{\phi_i},\sigma_{\phi_i}$ parametrize a Guassian distribution: $s^\prime = \mathcal{M}_{\phi_i}(s,a) \sim \mathcal{N}(\mu_{\phi_i}(s,a),\sigma_{\phi_i}(s,a))$. While training models based on samples from environment, we calculate bias averaged over the observed state-action pair $(S,A)$ for each model:
\begin{equation}
    \mathrm{bias}(\phi_i) = \underset{S,A\sim \pi,\mathcal{P}}{\mathbb{E}}\|\mathcal{M}_{\phi_i}(S,A)-\mathcal{P}(S,A)\|
\end{equation}
which formulates the distance of next states in model $\mathcal{M}_{\phi_i}$ and in environment $\mathcal{P}$, where $\|\cdot\|$ is a distance function on state space $\mathcal{S}$.

Then we select models from the model ensemble uniformly at random, sort them in ascending order by the calculated bias and retain a dropout subset with smaller model bias: $\mathcal{M}^\beta = \{\mathcal{M}_{\phi_1},\mathcal{M}_{\phi_2},\ldots,\mathcal{M}_{\phi_{N_{1-\beta}}}\}$, i.e., $\mathcal{M}^\beta=\left\{\mathcal{M}_\phi\mid \phi\in\Phi_\beta\right\}$, where $\Phi_\beta=\left\{\phi_i\mid \mathrm{bias}(\phi_i)\leq\mathrm{bias}(\phi_{N_{1-\beta}}),\phi_i\in\Phi\right\}$ and $N_{1-\beta}$ is the max integer in the ascending order index $\left\{1,2,\ldots,N_{1-\beta}\right\}$ after we dropout the $\beta$-percentile subset with large bias. 

\subsection{Theoretical Analysis of MBDP}\label{sec:guarantee}

We now give theoretical guarantees for the robustness and sample efficiency of the MBDP algorithm. \textbf{All the proofs of this section are detailed in Appendix A}.

\subsubsection{Guarantee of Robustness}

We define the robustness as the expected performance in a perturbed environment. Consider a perturbed transition matrix $\hat{\mathcal{P}}=\mathcal{P}_t \circ \delta_t$, where $\delta_t\in\mathbb{R}^{\mathcal{S}\times\mathcal{A}\times\mathcal{S}}$ is a multiplicative probability perturbation and $\circ$ is the Hadamard Product. Recall the definition of $\mathrm{CVaR}(\cdot)$ in equation (\ref{eq:def-cvar}), now we propose following theorem to provide guarantee of robustness for MBDP algorithm.


\begin{theorem}\label{the:robustness}
It holds
\begin{equation}\label{eq:return-cvar-rob}
    V_\alpha^{\pi,\mathcal{M}} = -\mathrm{CVaR}_\alpha(-V^{\pi,\mathcal{M}})=\sup\limits_{\Delta_\alpha}\mathbb{E}_{\hat{\mathcal{P}}}[V^{\pi,\mathcal{M}}]
\end{equation}
given the constraint set of perturbation
\begin{equation}\label{eq:supp-perturbation}
    \Delta_\alpha \triangleq \left\{\delta_i\middle\vert\prod_{i=1}^{T}\delta_i(s_i\mid s_{i-1},a_{i-1})\leq \frac{1}{\alpha}, \forall s_i\in\mathcal{S}, a_i\in\mathcal{A},\alpha\in(0,1)\right\}
\end{equation}

\end{theorem}

Since $\sup_{\Delta_\alpha}\mathbb{E}_{\hat{P}}[V^{\pi,\mathcal{M}}]$ means optimizing the expected performance in a perturbed environment, which is exactly our definition of robustness, then Theorem \ref{the:robustness} can be interpreted as an equivalence between optimizing robustness and the expected return under rollout-dropout, i.e., $V_\alpha^{\pi,\mathcal{M}}$.



\subsubsection{Guarantee of Efficiency}\label{sec:gua-effi}





We first propose Lemma \ref{lem:beta-drop-bound} to prove that the expected return with only rollout-dropout mechanism, compared to the expected return when it is deployed in the environment $\mathcal{P}$, has a discrepancy bound.

\begin{lemma}\label{lem:beta-drop-bound}

Suppose $\mathrm{R_{m}}$ is the supremum of reward function $r(s,a)$, i.e., $\mathrm{R_{m}}=\underset{s\in\mathcal{S},a\in\mathcal{A}}{\sup}r(s,a)$, the expected return of dropout batch rollouts with individual model $\mathcal{M}_\phi$ has a discrepancy bound:

\begin{equation}
    |{V}_\alpha^{\pi, \mathcal{M}_{\phi}} - {V}^{\pi,\mathcal{M}_{\phi}}| \leq \frac{\alpha(1+\alpha)}{(1-\alpha)(1-\gamma)}\mathrm{R_{m}} \triangleq \epsilon_\alpha
\label{eq:eps-beta}
\end{equation}

\end{lemma}


While Lemma \ref{lem:beta-drop-bound} only provides a guarantee for the performance of rollout-dropout mechanism, we now propose Theorem \ref{the:MBDP-bound} to prove that the expected return of policy derived by model dropout together with rollout-dropout, i.e., our MBDP algorithm, compared to the expected return when it is deployed in the environment $\mathcal{P}$, has a discrepancy bound.

\begin{theorem}\label{the:MBDP-bound}

Suppose $K\geq 0$ is a constant. The expected return of MBDP algorithm, i.e., ${V}_\alpha^{\pi, \mathcal{M}^\beta}$, compared to the expected return when it is deployed in the environment $\mathcal{P}$, i.e., ${V}^{\pi, \mathcal{P}}$, has a discrepancy bound:
\begin{equation}
\left|{V}_\alpha^{\pi, \mathcal{M}^\beta}-{V}^{\pi, \mathcal{P}}\right|\leq D_{\alpha,\beta}(\mathcal{M})
\end{equation}
where
\begin{equation}\label{eq:MBDP-bound}
D_{\alpha,\beta}(\mathcal{M})\triangleq\frac{(1-\beta)\gamma K}{1-\gamma}\epsilon_{\mathcal{M}}+\frac{\alpha(1+\alpha)(1-\beta)}{(1-\alpha)(1-\gamma)}\mathrm{R}_m
\end{equation}
and
\begin{equation}\label{def:delta-M}
\epsilon_{\mathcal{M}}\triangleq\underset{\phi\in\Phi}{\mathbb{E}}\left[\underset{s,a\sim \pi,\mathcal{P}}{\mathbb{E}}\left[\left\|\mathcal{M}_\phi(s, a)-\mathcal{P}(s, a)\right\|\right]\right]
\end{equation}

\end{theorem}


Since MBDP algorithm is an extension of the Dyna-style algorithm \cite{Sutton1991DynaReacting}: a series of model-based reinforcement learning methods which jointly optimize the policy and transition model, it can be written in a general pattern as below:
\begin{equation}
    \pi_{k+1}, \mathcal{M}^\beta_{k+1}=\underset{\pi_k, \mathcal{M}_k^\beta}{\arg\max}\left[{V}^{\pi_k, \mathcal{M}_k^\beta}-D_{\alpha,\beta}(\mathcal{M}^\beta_k)\right]
\label{eq:algo-equation}
\end{equation}
where $\pi_k$ denotes the updated policy in $k$-th iteration and $\mathcal{M}^\beta_{k}$ denotes the updated dropout model ensemble in $k$-th iteration. In this setting, we can show that, performance of the policy derived by our MBDP algorithm, is approximatively monotonically increasing when deploying in the real environment $\mathcal{P}$, with ability to robustly jump out of local optimum.

\begin{proposition}
The expected return of policy derived by general algorithm pattern (\ref{eq:algo-equation}), is approximatively monotonically increasing when deploying in the real environment $\mathcal{P}$, i.e.
\begin{equation}
    {V}^{\pi_{k+1}, \mathcal{P}}\geq {V}^{\pi_{k}, \mathcal{P}} + (\epsilon_{k+1} - \epsilon_\alpha) \triangleq {V}^{\pi_{k}, \mathcal{P}} + \eta
\end{equation}
where $\epsilon_\alpha$ is defined in (\ref{eq:eps-beta}) and $\epsilon_{k+1}$ is the update residual:
\begin{equation}
    \epsilon_{k+1} \triangleq {V}^{\pi_{k+1}, \mathcal{P}} - \left[{V}_{\alpha}^{\pi_{k+1}, \mathcal{M}_{k+1}^\beta} - D_{\alpha,\beta}(\mathcal{M}_{k+1}^\beta)\right]
\end{equation}
\label{prop:performance}
\end{proposition}

Intuitively, proposition \ref{prop:performance} shows that under the control of reasonable parameters $\alpha$ and $\beta$, $\epsilon_{k+1}$ is often a large update value in the early learning stage, while $\epsilon_\alpha$ as an error bound is a fixed small value. Thus $\eta=\epsilon_{k+1}-\epsilon_\alpha$ is a value greater than $0$ most of the time in the early learning stage, which can guarantee ${V}^{\pi_{k+1}, \mathcal{P}}\geq {V}^{\pi_{k}, \mathcal{P}} + 0$. In the late stage near convergence, the update becomes slow and $\epsilon_{k+1}$ may be smaller than $\epsilon_\alpha$, which leads to the possibility that $V_{k+1}$ is smaller than $V_k$. This makes the update process try some other convergence direction, providing an opportunity to jump out of the local optimum. We empirically verify this claim in Appendix C.

\subsubsection{Flexible control of robustness and efficiency}\label{sec:flex-control}

According to Theorem \ref{the:robustness}, rollout-dropout improves robustness, and the larger $\alpha$ is, the more robustness is improved. Conversely, the smaller $\alpha$ is, the worse the robustness will be. For model-dropout, it is obvious that when $\beta$ is larger, it means that the more models we will be dropped, and the more likely the model is to overfit the environment, so the less robust it is. Conversely, when $\beta$ is less, the model ensemble has better robustness in simulating complex environments, and the robustness is better at this point.

Turning to the efficiency. Note that the bound in equation (\ref{eq:MBDP-bound}) i.e., $D_{\alpha,\beta}(\mathcal{M})$, is in positive ratio with $\alpha$ and inverse ratio with $\beta$. This means that as $\alpha$ increases or $\beta$ decreases, this bound expands, causing the accuracy of the algorithm to decrease and the algorithm to take longer to converge, thus making it less efficient. Conversely, when $\alpha$ decreases or $\beta$ increases, the efficiency increases.

With the analysis above, it suggests that MBDP can provide a flexible control mechanism to meet different demands of robustness and efficiency by tuning two corresponding dropout ratios. This conclusion can be summarized as follows and we also empirically verify it in section \ref{sec:experiments}.

\begin{itemize}
    \item \textbf{To get balanced efficiency and robustness}: set $\alpha$ and $\beta$ both to a moderate value
    \item \textbf{To get better robustness}: set $\alpha$ to a larger value and $\beta$ to a smaller value.
    \item \textbf{To get better efficiency}: set $\alpha$ to a smaller value and $\beta$ to a larger value.
\end{itemize}

\section{Experiments}\label{sec:experiments}

Our experiments aim to answer the following questions: 

\begin{itemize}
    \item How does MBDP perform on benchmark reinforcement learning tasks compared to state-of-the-art model-based and model-free RL methods?
    \item Can MBDP find a balance between robustness and benefits?
    \item How does the robustness and efficiency of MBDP change by tuning parameters $\alpha$ and $\beta$?
\end{itemize}

To answer the posed questions, we need to understand how well our method compares to state-of-the-art model-based and model-free methods and how our design choices affect performance. We evaluate our approach on four continuous control benchmark tasks in the Mujoco simulator \cite{todorov2012mujoco}: Hopper, Walker, HalfCheetah, and Ant. We also need to perform the ablation study by removing the dropout modules from our algorithm. Finally, a separate analysis of the hyperparameters ($\alpha$ and $\beta$) is also needed. A depiction of the environments and a detailed description of the experimental setup can be found in Appendix B.

\subsection{Comparison with State-of-the-Arts}

In this subsection, we compare our MBDP algorithm with state-of-the-art model-free and model-based reinforcement learning algorithms in terms of sample complexity and performance. Specifically, we compare against SAC \cite{haarnoja2018soft}, which is the state-of-the-art model-free method and establishes a widely accepted baseline. For model-based methods, we compare against MBPO \cite{janner2019trust}, which uses short-horizon model-based rollouts started from samples in the real environment; STEVE \cite{buckman2018sample}, which dynamically incorporates data from rollouts into value estimation rather than policy learning; and SLBO \cite{Luo2019AlgorithmicGuarantees}, a model-based algorithm with performance guarantees. For our MBDP algorithm, we choose $\alpha=0.2$ and $\beta=0.2$ as hyperparameter setting.

\begin{figure*}
  \centering
  \includegraphics[width=\textwidth]{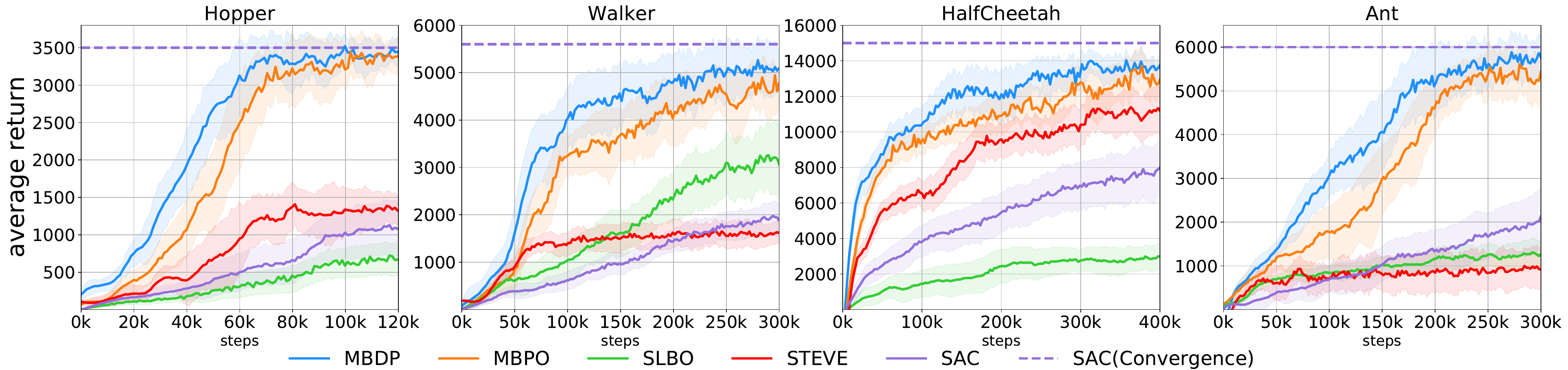}
  \caption{Learning curves of our MBDP algorithm and four baselines on different continuous control environments. Solid curves indicate the mean of all trials with 5 different seeds. Shaded regions correspond to standard deviation among trials. Each trial is evaluated every 1000 steps. The dashed reference lines are the asymptotic performance of SAC algorithm. These results show that our MBDP method learns faster and has better asymptotic performance and sample efficiency than existing model-based algorithms.}
  \label{fig:performance}
\end{figure*}

Figure \ref{fig:performance} shows the learning curves for all methods, along with asymptotic performance of the model-free SAC algorithm which do not converge in the region shown. The results highlight the strength of MBDP in terms of performance and sample complexity. In all the Mujoco simulator environments, our MBDP method learns faster and has better efficiency than existing model-based algorithms, which empirically demonstrates the advantage of Dropout Planning.

\subsection{Analysis of Robustness}

\begin{figure*}
  \centering
  \includegraphics[width=\textwidth]{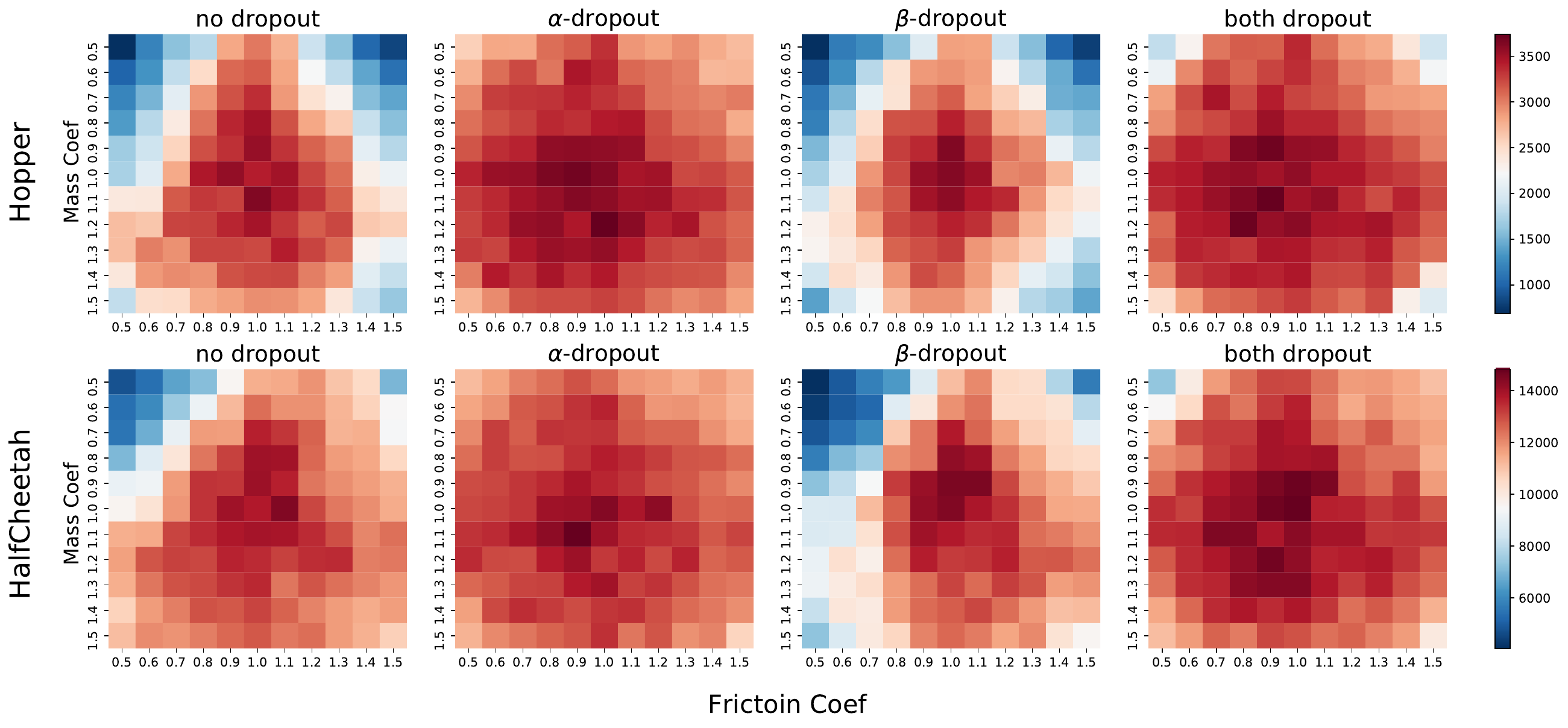}
  \caption{The robustness performance is depicted as heat maps for various environment settings. Each heat map represents a set of experiments, and each square in the heat map represents the average return value in one experiment. The closer the color to red (hotter) means the higher the value, the better the algorithm is trained in that environment, and vice versa. The four different algorithms in the figure are no dropout ($\alpha=0, \beta=0$), rollout-dropout only ($\alpha$-dropout: $\alpha=0.2, \beta=0$), model-dropout only ($\beta$-dropout: $\alpha=0, \beta=0.2$), and both dropouts ($\alpha=0.2, \beta=0.2$). Each experiment in the Hopper environment stops after 300,000 steps, and each experiment in the HalfCheetah environment stops after 600,000 steps.}
  \label{fig:robustness-heatmap}
\end{figure*}

Aiming to evaluate the robustness of our MBDP algorithm by testing policies on different environment settings (i.e., different combinations of physical parameters) without any adaption, we define ranges of mass and friction coefficients as follows: $0.5\leq C_\mathrm{mass}\leq 1.5$ and $0.5\leq C_\mathrm{friction}\leq 1.5$, and modify the environments by scaling the torso mass with coefficient $C_\mathrm{mass}$ and the friction of every geom with coefficient $C_\mathrm{friction}$. 

We compare the original MBDP algorithm with the $\alpha$-dropout variation ($\alpha=0.2, \beta=0$) which keeps only the rollout-dropout, the $\beta$-dropout variation ($\alpha=0, \beta=0.2$) which keeps only the model-dropout, and the no-dropout variation ($\alpha=0, \beta=0$) which removes both dropouts. This experiment is conducted in the modified environments mentioned above. The results are presented in Figure \ref{fig:robustness-heatmap} in the form of heat maps, each square of a heat map represents the average return value that the algorithm can achieve after training in each modified environment. The closer the color to red (hotter) means the higher the value, the better the algorithm is trained in that environment, and vice versa. Obviously, if the algorithm can only achieve good training results in the central region and inadequate results in the region far from the center, it means that the algorithm is more sensitive to perturbation in environments and thus less robust.

Based on the results, we can see that the $\alpha$-dropout using only the rollout-dropout can improve the robustness of the algorithm, while the $\beta$-dropout using only the model-dropout will slightly weaken the robustness, and the combination of both dropouts, i.e., the MBDP algorithm, achieves robustness close to that of $\alpha$-dropout.

\subsection{Ablation Study}

In this section, we investigate the sensitivity of MBDP algorithm to the hyperparameter $\alpha, \beta$. We conduct two sets of experiments in both Hopper and HalfCheetah environments: (1) fix $\beta$ and change $\alpha$ ($\alpha\in[0,0.5], \beta = 0.2$); (2) fix $\alpha$ and change $\beta$ ($\beta\in[0,0.5], \alpha = 0.2$).

\begin{figure}
  \centering
  \includegraphics[width=\textwidth]{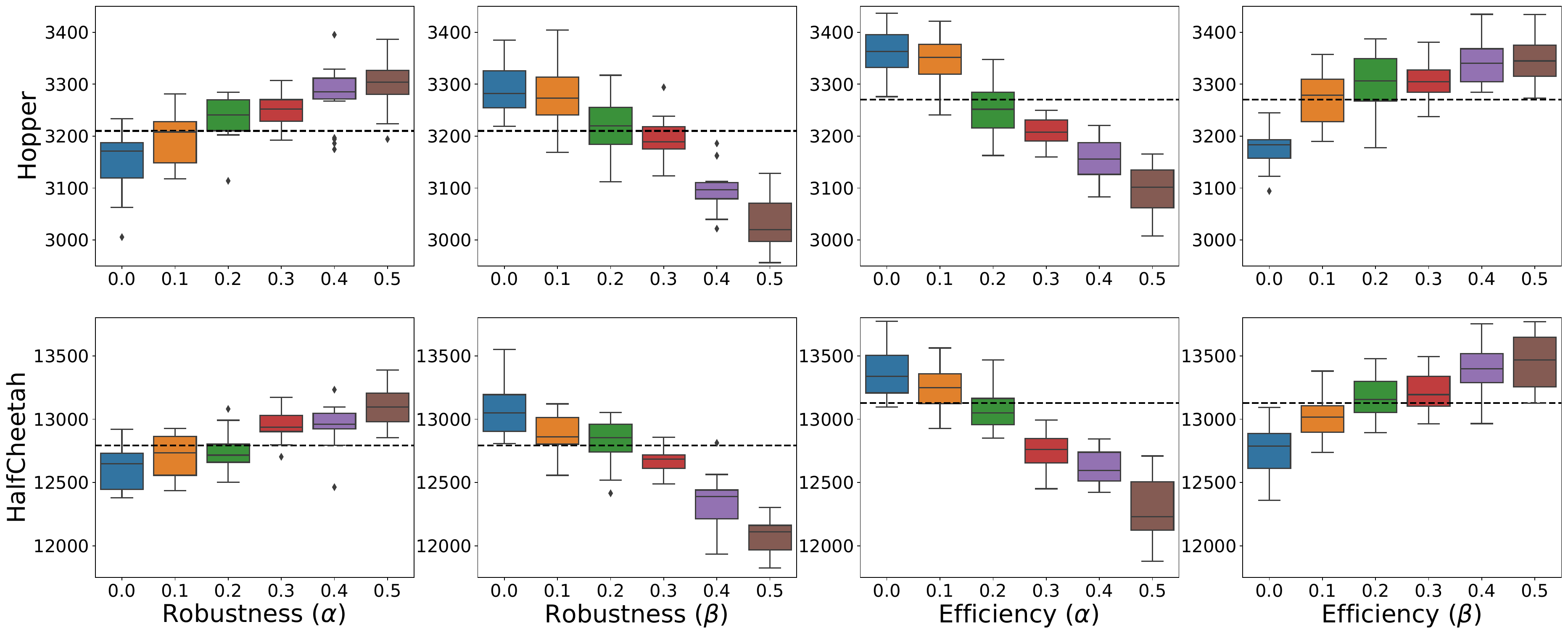}
  \caption{The horizontal axis represents the different values of $\alpha,\beta$. The vertical axis is the metric for evaluating the robustness or efficiency. The horizontal dashed line is the baseline case with both rollout-dropout and model-dropout removed ($\alpha=\beta=0$). 120k steps are trained for each experiment in the Hopper environment, and 400k steps are trained for each experiment in the HalfCheetah environment. Each box plot corresponds to 10 different random seeds.}
  \label{fig:hyper-performance}
\end{figure}

The experimental results are shown in Figure \ref{fig:hyper-performance}. The first row corresponds to experiments in the Hopper environment and the second row corresponds to experiments in the HalfCheetah environment. Columns 1 and 2 correspond to the experiments conducted in the perturbed Mujoco environment with modified environment settings. We construct a total of $2\times 2=4$ different perturbed environments ($C_\mathrm{mass} = 0.8, 1.2, C_\mathrm{friction} = 0.8, 1.2$), and calculate the average of the return values after training a fixed number of steps (Hopper: 120k steps, HalfCheetah: 400k steps) in each of the four environments. The higher this average value represents the algorithm can achieve better overall performance in multiple perturbed environments, implying better robustness. Therefore, this metric can be used to evaluate the robustness of different $\alpha,\beta$. Columns 3 and 4 are the return values obtained after a fixed number of steps (Hopper: 120k steps, HalfCheetah: 400k steps) for experiments conducted in the standard Mujoco environment without any modification, which are used to evaluate the efficiency of the algorithm for different values of $\alpha,\beta$. Each box plot corresponds to 10 different random seeds.

Observing the experimental results, we can find that robustness shows a positive relationship with $\alpha$ and an inverse relationship with $\beta$; efficiency shows an inverse relationship with $\alpha$ and a positive relationship with $\beta$. This result verifies our conclusion in Section \ref{sec:flex-control}. In addition, we use horizontal dashed lines in Figure \ref{fig:hyper-performance} to indicate the baseline with rollout-dropout and model-dropout removed ($\alpha=\beta=0$). It can be seen that when $\alpha\in[0.1,0.2],\beta\in[0.1,0.2]$, the robustness and efficiency of the algorithm can both exceed the baseline. Therefore, when $\alpha,\beta$ is adjusted to a reasonable range of values, we can simultaneously improve the robustness and efficiency.

\section{Conclusions and Future Work}
In this paper, we propose the MBDP algorithm to address the dilemma of robustness and sample efficiency. Specifically, MBDP drops some overvalued imaginary samples through the rollout-dropout mechanism to focus on the bad samples for the purpose of improving robustness, while the model-dropout mechanism is designed to enhance the sample efficiency by only using accurate models. Both theoretical analysis and experiment results verify our claims that 1) MBDP algorithm can provide policies with competitive robustness while achieving state-of-the-art performance; 2) we empirically find that there is a seesaw phenomenon between robustness and efficiency, that is, the growth of one will cause a slight decline of the other; 3) we can get policies with different types of performance and robustness by tuning the hyperparameters $\alpha$ and $\beta$, ensuring that our algorithm is capable of performing well in a wide range of tasks.

Our future work will incorporate more domain knowledge of robust control to further enhance robustness. We also plan to transfer the design of Double Dropout Planning as a more general module that can be easily embedded in more model-based RL algorithms and validate the effectiveness of Double Dropout Planning in real-world scenarios. Besides, relevant researches in the field of meta learning and transfer learning may inspire us to further optimize the design and training procedure of the predictive models. Finally, we can use more powerful function approximators to model the environment.

\bibliographystyle{unsrt}
\bibliography{MBDP}

\newpage

\appendix

\section{Proofs}

In Appendix A, we will provide proofs for Theorem \ref{the:robustness}, Lemma \ref{lem:beta-drop-bound}, Theorem \ref{the:MBDP-bound} ,and Proposition \ref{prop:performance}. Note that the numbering and citations in the appendices are referenced from the main manuscript.

\subsection{Proof of Theorem \ref{the:robustness}}

\begin{proof}
Recall the definition of $\mathrm{CVaR}$ (\ref{eq:def-cvar}) and $V_\alpha^{\pi,\mathcal{M}}$
(\ref{def:eta-beta}), we need to take the negative value of rewards to represent the loss in the sense of CVaR. Then we have that,

\begin{align*}
\mathrm{CVaR}_\alpha(-V^{\pi,\mathcal{M}}) &= \mathbb{E}\left[-V^{\pi,\mathcal{M}}\mid -V^{\pi,\mathcal{M}} \geq \mathrm{VaR}_\alpha(-V^{\pi,\mathcal{M}})\right]\\
&=\mathbb{E}\left[-{\sum}_{\mathcal{B}^{\pi,\mathcal{M}}}\left[\gamma^t r(s_t,a_t)\right]\mid -{\sum}_{\mathcal{B}^{\pi,\mathcal{M}}}\left[\gamma^t r(s_t,a_t)\right] \geq \mathrm{VaR}_\alpha(-V^{\pi,\mathcal{M}})\right]\\
&=-\mathbb{E}\left[{\sum}_{\mathcal{B}^{\pi,\mathcal{M}}}\left[\gamma^t r(s_t,a_t)\right]\mid {\sum}_{\mathcal{B}^{\pi,\mathcal{M}}}\left[\gamma^t r(s_t,a_t)\right] \leq -\mathrm{VaR}_\alpha(-V^{\pi,\mathcal{M}})\right]\\
\end{align*}

Obviously, the condition of ${\sum}_{\mathcal{B}^{\pi,\mathcal{M}}}\left[\gamma^t r(s_t,a_t)\right] \leq -\mathrm{VaR}_\alpha(-V^{\pi,\mathcal{M}})$ in the above equation exactly meets our definition of $\mathcal{B}_\alpha^{\pi,\mathcal{M}}$, that is, eqaution (\ref{def:batch-alpha-beta}). Then we can prove the first part of Theorem \ref{the:robustness}

\begin{equation}
-\mathrm{CVaR}_\alpha(-V^{\pi,\mathcal{M}}) = \mathbb{E}\left[{\sum}_{\mathcal{B}_\alpha^{\pi,\mathcal{M}}}\left[\gamma^t r(s_t,a_t)\right]\right] = V_\alpha^{\pi,\mathcal{M}}
\end{equation}

Considering $\mathbb{E}_{\hat{\mathcal{P}}}[-V^{\pi,\mathcal{M}}]$, recall the definition of $\hat{\mathcal{P}}$, we have that

\begin{align*}
    \mathbb{E}_{\hat{\mathcal{P}}}[V^{\pi,\mathcal{M}}] &= -\mathbb{E}_{\hat{\mathcal{P}}}[-V^{\pi,\mathcal{M}}]\\
    &= -\sum_{(s_0,\ldots,s_T)\in\mathcal{S}^{T+1}}\mathcal{P}_0(s_0)\delta_0(s_0)\prod_{t=1}^{T}\mathcal{P}_t(s_t\mid s_{t-1})\delta_t(s_t\mid s_{t-1})\cdot (-V^{\pi,\mathcal{M}})\\
    &= \sum_{(s_0,\ldots,s_T)\in\mathcal{S}^{T+1}}\mathcal{P}(s_0,\ldots,s_T)\delta_0(s_0)\prod_{t=1}^{T}\delta_t(x_t\mid x_{t-1})\cdot V^{\pi,\mathcal{M}}\\
    &\triangleq \sum_{(s_0,\ldots,s_T)\in\mathcal{S}^{T+1}}\mathcal{P}(s_0,\ldots,s_T)\delta(s_0,\ldots,s_T)\cdot V^{\pi,\mathcal{M}}
\end{align*}

Since $\delta$ is the random perturbation to the environment as we defined, it's intuitive that

\begin{equation}\label{eq:delta-exp}
    \mathbb{E}\left[\delta(s_0,\ldots,s_T)\right] = \sum_{(s_0,\ldots,s_T)\in\mathcal{S}^{T+1}}\mathcal{P}(s_0,\ldots,s_T)\delta(s_0,\ldots,s_T) = 1
\end{equation}

Recall the definition of $\Delta_\alpha$ in (\ref{eq:supp-perturbation}), we can prove the second part of Theorem \ref{the:robustness}

\begin{align}
    \sup\limits_{\Delta_\alpha}\mathbb{E}_{\hat{\mathcal{P}}}[V^{\pi,\mathcal{M}}] &= \sup\limits_{\delta(s_0,\ldots,s_T)\leq\frac{1}{\alpha}}\sum_{(s_0,\ldots,s_T)\in\mathcal{S}^{T+1}}\mathcal{P}(s_0,\ldots,s_T)\delta(s_0,\ldots,s_T)\cdot V^{\pi,\mathcal{M}}\notag\\
    &=-\mathrm{CVaR}_\alpha(-V^{\pi,\mathcal{M}})\label{proof:cvar-eq}
\end{align}

The last equation (\ref{proof:cvar-eq}) is obtained by equation (\ref{eq:delta-exp}) and the Representation Theorem [22] for CVaR.

\end{proof}

\subsection{Proof of Lemma \ref{lem:beta-drop-bound}}

To prove Lemma \ref{lem:beta-drop-bound}, we need to introduce two useful lemmas.

\begin{lemma}\label{lem:proof-for-lem41}

Define

\begin{equation}\label{def:G-sa}
G^{\pi,\mathcal{M}}(s,a)=\underset{\hat{s}'\sim\mathcal{M}(\cdot|s,a)}{\mathbb{E}}{{V}^{\pi,\mathcal{M}}}(\hat{s}') - \underset{s'\sim\mathcal{P}(\cdot|s,a)}{\mathbb{E}}{{V}^{\pi,\mathcal{M}}}(s')
\end{equation}
For any policy $\pi$ and dynamical models $\mathcal{M},\mathcal{M}'$, we have that

\begin{equation}
{V}^{\pi,\mathcal{M}'} - {V}^{\pi,\mathcal{M}} = \frac{\gamma}{1-\gamma}\underset{S,A\sim\pi,\mathcal{M}}{\mathbb{E}}\left[G^{\pi,\mathcal{M}'}(S,A)\right]
\end{equation}

\end{lemma}

Lemma \ref{lem:proof-for-lem41} is a directly cited theorem in existing work (Lemma 4.3 in [31]), we make some modifications to fit our subsequent conclusions. With the above lemma, we first propose Lemma \ref{lem:mb-bound}.

\begin{lemma}\label{lem:mb-bound}
Suppose the expected return for model-based methods ${V}^{\pi,\mathcal{M}}$ is Lipschitz continuous on the state space $\mathcal{S}$, $K$ is the Lipschitz constant, $\mathcal{P}$ is the transition distribution of environment, then
\begin{equation}
\left|{V}^{\pi, \mathcal{M}}-{V}^{\pi, \mathcal{P}}\right| \leq\frac{\gamma}{1-\gamma}K\cdot\mathrm{bias}
\end{equation}
where
\begin{equation}
\mathrm{bias} \triangleq \underset{s,a\sim \pi,\mathcal{P}}{\mathbb{E}}\left\|\mathcal{M}(s, a)-\mathcal{P}(s, a)\right\|
\end{equation}

\label{theo:mb-bound}
\end{lemma}

In Lemma \ref{lem:mb-bound}, we make the assumption that the expected return ${V}^{\pi,\mathcal{M}}(s)$ on the estimated model $\mathcal{M}$ is \textit{Lipschitz continuous} w.r.t any norm $\|\cdot\|$, i.e.

\begin{equation}\label{assum:lip}
    \left|{V}^{\pi,\mathcal{M}}(s)-{V}^{\pi,\mathcal{M}}(s^{\prime})\right| \leq K\left\|s-s^{\prime}\right\|, \forall s, s^{\prime} \in \mathcal{S}
\end{equation}
where $K\in \mathbb{R}^+$ is a \textit{Lipschitz constant}. This assumption means that the closer states should give the closer value estimation, which should hold in most scenarios.

\begin{proof}

By definition of $G^{\pi,\mathcal{M}}(s,a)$ in (\ref{def:G-sa}) and Assumption (\ref{assum:lip}), i.e., ${V}^{\pi,\mathcal{M}}(s)$ is \textit{Lipschitz continuous}, we have that

\begin{equation}\label{eq:G-leq}
|G^{\pi,\mathcal{M}}(s,a)|\leq K\|\mathcal{M}(s,a)-\mathcal{P}(s,a)\|
\end{equation}
Then, we can show that

\begin{align*}
\left|{V}^{\pi, \mathcal{M}}-{V}^{\pi, \mathcal{P}}\right| &= \frac{\gamma}{1-\gamma}\left|\underset{s,a\sim\pi,\mathcal{P}}{\mathbb{E}}\left[G^{\pi,\mathcal{M}}(s,a)\right]\right| \tag{By Lemma \ref{lem:proof-for-lem41}}\\
&\leq \frac{\gamma}{1-\gamma}\underset{s,a\sim\pi,\mathcal{P}}{\mathbb{E}}\left[\left|G^{\pi,\mathcal{M}}(s,a)\right|\right] \tag{By Triangle Inequality}\\
&\leq \frac{\gamma}{1-\gamma}\underset{s,a\sim\pi,\mathcal{P}}{\mathbb{E}}K\|\mathcal{M}(s,a)-\mathcal{P}(s,a)\| \tag{By equation (\ref{eq:G-leq})}\\
&= \frac{\gamma}{1-\gamma}K\cdot\underset{s,a\sim\pi,\mathcal{P}}{\mathbb{E}}\|\mathcal{M}(s,a)-\mathcal{P}(s,a)\|\\
&\triangleq \frac{\gamma}{1-\gamma}K\cdot\mathrm{bias}
\end{align*}

\end{proof}

\newpage

Now we prove Lemma \ref{lem:beta-drop-bound}.

\begin{proof}

For two disjoint sets $A$ and $B$, i.e., $A\cap B=\varnothing$, there are the following property

\begin{equation}
    \mathbb{E}_{A\cup B}[X] = \mathbb{E}_A[X]\mathrm{P}(A)+\mathbb{E}_B[X]\mathrm{P}(B)
\end{equation}

By this property,

\begin{align*}
    &\mathbb{E}\left[{\sum}_{s_0\in\mathcal{S},\{a_0,s_1,\ldots\} \sim\pi,\mathcal{M}_\phi}\left[\gamma^t r(s_t,a_t)\right]\right] \\
    &= (1-\alpha)\cdot \mathbb{E}\left[{\sum}_{\{s_0,a_0,\ldots\} \sim\mathcal{B}_\alpha^{\pi,\mathcal{M}_\phi}}\left[\gamma^t r(s_t,a_t)\right]\right] + \alpha\cdot\mathbb{E}\left[{\sum}_{\{s_0,a_0,\ldots\} \not\sim\mathcal{B}_\alpha^{\pi,\mathcal{M}_\phi}}\left[\gamma^t r(s_t,a_t)\right]\right]
\end{align*}

Recall the definition (\ref{def:eta-s}), (\ref{def:eta-expectation}) and (\ref{def:eta-beta}), we have that

\begin{align}
{V}^{\pi,\mathcal{M}_\phi}_\alpha&=\mathbb{E}\left[{\sum}_{\{s_0,a_0,\ldots\} \sim\mathcal{B}_\alpha^{\pi,\mathcal{M}_\phi}}\left[\gamma^t r(s_t,a_t)\right]\right]\notag\\
&=\frac{1}{1-\alpha}\mathbb{E}\left[{\sum}_{s_0\in\mathcal{S},\{a_0,s_1,\ldots\} \sim\pi,\mathcal{M}_\phi}\left[\gamma^t r(s_t,a_t)\right]\right]-\frac{\alpha}{1-\alpha}\mathbb{E}\left[{\sum}_{\{s_0,a_0,\ldots\} \not\sim\mathcal{B}_\alpha^{\pi,\mathcal{M}_\phi}}\left[\gamma^t r(s_t,a_t)\right]\right]\notag\\
&=\frac{1}{1-\alpha}\underset{s\in{\mathcal{S}}}{\mathbb{E}}\left[{V}^{\pi,\mathcal{M}_\phi}(s)\right]-\frac{\alpha}{1-\alpha}\mathbb{E}\left[{\sum}_{\tau \not\sim\mathcal{B}_\alpha^{\pi,\mathcal{M}_\phi}}\left[\gamma^t r(s_t,a_t)\right]\right]\notag\\
&=\frac{1}{1-\alpha}{V}^{\pi,\mathcal{M}_\phi}-\frac{\alpha}{1-\alpha}\mathbb{E}\left[{\sum}_{\tau \not\sim\mathcal{B}_\alpha^{\pi,\mathcal{M}_\phi}}\left[\gamma^t r(s_t,a_t)\right]\right]\label{proof:lem42p1}
\end{align}


Where $\tau\triangleq\{s_0,a_0,\ldots\}$. Recall the definition (\ref{def:batch-alpha-beta}) and  $\mathrm{R}_{m}=\underset{s\in\mathcal{S},a\in\mathcal{A}}{\sup}r(s,a)$, we have

\begin{align*}
\mathbb{E}\left[{\sum}_{\tau \not\sim\mathcal{B}_\alpha^{\pi,\mathcal{M}_\phi}}\left[\gamma^t r(s_t,a_t)\right]\right] &\leq\int_{\tau\not\sim{\mathcal{B}_\alpha^{\pi,\mathcal{M}_\phi}}}\left[\sum_{t=0}^\infty\gamma^t \mathrm{R}_m\right]p(\tau)\mathrm{d}\tau\\
&=\left[\sum_{t=0}^\infty\gamma^t\right]\mathrm{R}_m\int_{\tau\not\sim{\mathcal{B}_\alpha^{\pi,\mathcal{M}_\phi}}}p(\tau)\mathrm{d}\tau\\
&= \frac{1}{1-\gamma}\mathrm{R}_m \int_{\tau\not\sim{\mathcal{B}_\alpha^{\pi,\mathcal{M}_\phi}}}p(\tau)\mathrm{d}\tau\\
&=\frac{\alpha}{1-\gamma}\mathrm{R}_m \tag{By definition of $\mathcal{B}_\alpha^{\pi,\mathcal{M}_\phi}$}\\
\end{align*}

Similarly,

\begin{align*}
V^{\pi,\mathcal{M}_\phi} = \mathbb{E}\left[{\sum}_{\tau \sim\mathcal{B}^{\pi,\mathcal{M}_\phi}}\left[\gamma^t r(s_t,a_t)\right]\right] &\leq\int_{\tau \sim\mathcal{B}^{\pi,\mathcal{M}_\phi}}\left[\sum_{t=0}^\infty\gamma^t \mathrm{R}_m\right]p(\tau)\mathrm{d}\tau\\
&=\left[\sum_{t=0}^\infty\gamma^t\right]\mathrm{R}_m\int_{\tau \sim\mathcal{B}^{\pi,\mathcal{M}_\phi}}p(\tau)\mathrm{d}\tau\\
&= \frac{1}{1-\gamma}\mathrm{R}_m \int_{\tau \sim\mathcal{B}^{\pi,\mathcal{M}_\phi}}p(\tau)\mathrm{d}\tau\\
&=\frac{1}{1-\gamma}\mathrm{R}_m\\
\end{align*}

Based on the above two inequalities and equation (\ref{proof:lem42p1}), we have that

\begin{align}
|{V}_\alpha^{\pi, \mathcal{M}_{\phi}} - {V}^{\pi,\mathcal{M}_{\phi}}| &=  \left|\frac{\alpha}{1-\alpha}V^{\pi,\mathcal{M}_\phi}-\frac{\alpha}{1-\alpha}\mathbb{E}\left[{\sum}_{\tau \not\sim\mathcal{B}_\alpha^{\pi,\mathcal{M}_\phi}}\left[\gamma^t r(s_t,a_t)\right]\right]\right| \notag\\
&\leq\frac{\alpha}{1-\alpha}\left(\left|V^{\pi,\mathcal{M}_\phi}\right|+\left|\mathbb{E}\left[{\sum}_{\tau \not\sim\mathcal{B}_\alpha^{\pi,\mathcal{M}_\phi}}\left[\gamma^t r(s_t,a_t)\right]\right]\right|\right)\notag\\
&\leq \frac{\alpha}{1-\alpha}\left(\frac{1}{1-\gamma}\mathrm{R}_m+\frac{\alpha}{1-\gamma}\mathrm{R}_m\right) \notag\\
&=\frac{\alpha(1+\alpha)}{(1-\alpha)(1-\gamma)}\mathrm{R_{m}} \label{eq:lem42}
\end{align}

\end{proof}

\subsection{Proof of Theorem \ref{the:MBDP-bound}}

\begin{proof}

With Lemma \ref{lem:mb-bound} and Lemma \ref{lem:beta-drop-bound}, we can show that

\begin{align}
\left|{V}_\alpha^{\pi, \mathcal{M}^\beta}-{V}^{\pi, \mathcal{P}}\right|&=\left|\int_{\Phi_\beta}{V}_\alpha^{\pi, \mathcal{M}_{\phi}}p(\phi)\mathrm{d}\phi-{V}^{\pi, \mathcal{P}}\right| \notag\\
&=\left|\int_{\Phi_\beta}\left({V}_\alpha^{\pi, \mathcal{M}_{\phi}}-{V}^{\pi, \mathcal{P}}\right)p(\phi)\mathrm{d}\phi\right| \notag\\
&\leq\int_{\Phi_\beta}\left|{V}_\alpha^{\pi, \mathcal{M}_{\phi}}-{V}^{\pi, \mathcal{P}}\right|p(\phi)\mathrm{d}\phi \tag{By Triangle Inequality}\\
&\leq\int_{\Phi_\beta}\left(\left|{V}^{\pi, \mathcal{M}_{\phi}}-{V}^{\pi, \mathcal{P}}\right|+\left|{V}_\alpha^{\pi, \mathcal{M}_{\phi}} - {V}^{\pi,\mathcal{M}_{\phi}}\right|\right)p(\phi)\mathrm{d}\phi\quad \tag{By Lemma \ref{lem:beta-drop-bound}}\\
&= \int_{\Phi_\beta}\left|{V}^{\pi, \mathcal{M}_{\phi}}-{V}^{\pi, \mathcal{P}}\right|p(\phi)\mathrm{d}\phi+\int_{\Phi_\beta}\left|{V}_\alpha^{\pi, \mathcal{M}_{\phi}} - {V}^{\pi,\mathcal{M}_{\phi}}\right|p(\phi)\mathrm{d}\phi \label{proof:theo43p1}
\end{align}
For the first part of (\ref{proof:theo43p1}), let

$$
\epsilon_{\mathcal{M}}\triangleq\underset{\phi\in\Phi}{\mathbb{E}}\left[\underset{s,a\sim \pi,\mathcal{P}}{\mathbb{E}}\left[\left\|\mathcal{M}_\phi(s, a)-\mathcal{P}(s, a)\right\|\right]\right]
$$
denotes the general bias between any model $\mathcal{M}$ and environment transition $\mathcal{P}$, with Lemma \ref{lem:mb-bound}, we now get

\begin{align}
\int_{\Phi_\beta}\left|{V}^{\pi, \mathcal{M}_{\phi}}-{V}^{\pi, \mathcal{P}}\right|p(\phi)\mathrm{d}\phi &\leq \frac{\gamma K}{1-\gamma}\int_{\Phi_\beta}\underset{s,a\sim\pi,\mathcal{P}}{\mathbb{E}}\left[\left\|\mathcal{M}_\phi(s, a)-\mathcal{P}(s, a)\right\|\right]p(\phi)\mathrm{d}\phi \tag{By Lemma \ref{lem:mb-bound}}\\
&\leq \frac{\gamma K}{1-\gamma}\left|\epsilon_{\mathcal{M}}\right|\int_{\Phi_\beta}\left|p(\phi)\right|\mathrm{d}\phi\notag\\
&=\frac{(1-\beta)\gamma K}{1-\gamma}\epsilon_{\mathcal{M}} \label{proof:theo43p2}
\end{align}
For the second part of (\ref{proof:theo43p1}), by Lemma \ref{lem:beta-drop-bound}, we can show that

\begin{align}
\int_{\Phi_\beta}\left|{V}_\alpha^{\pi, \mathcal{M}_{\phi}} - {V}^{\pi,\mathcal{M}_{\phi}}\right|p(\phi)\mathrm{d}\phi&\leq \frac{\alpha(1+\alpha)}{(1-\alpha)(1-\gamma)}\mathrm{R}_m\int_{\Phi_\beta}p(\phi)\mathrm{d}\phi \tag{By Lemma \ref{lem:beta-drop-bound}}\\
&=\frac{\alpha(1+\alpha)(1-\beta)}{(1-\alpha)(1-\gamma)}\mathrm{R}_m \label{proof:theo43p3}
\end{align}
Go back to equation (\ref{proof:theo43p1}), it follows that

\begin{align*}
\left|{V}_\alpha^{\pi, \mathcal{M}^\beta}-{V}^{\pi, \mathcal{P}}\right| &\leq \int_{\Phi_\beta}\left|{V}^{\pi, \mathcal{M}_{\phi}}-{V}^{\pi, \mathcal{P}}\right|p(\phi)\mathrm{d}\phi+\int_{\Phi_\beta}\left|\epsilon_\alpha\right|p(\phi)\mathrm{d}\phi\\
&\leq \frac{(1-\beta)\gamma K}{1-\gamma}\epsilon_{\mathcal{M}}+\frac{\alpha(1+\alpha)(1-\beta)}{(1-\alpha)(1-\gamma)}\mathrm{R}_m \tag{By equation (\ref{proof:theo43p2}) and (\ref{proof:theo43p3})}\\
&\triangleq D_{\alpha,\beta}(\mathcal{M})
\end{align*}

\end{proof}

\subsection{Proof of Proposition \ref{prop:performance}}

\begin{proof}

With Theorem \ref{the:MBDP-bound}, i.e., $\left|{V}_\alpha^{\pi, \mathcal{M}^\beta}-{V}^{\pi, \mathcal{P}}\right|\leq D_{\alpha,\beta}(\mathcal{M})$, we have

\begin{equation}\label{eq:prop44p1}
{V}^{\pi_{k+1}, \mathcal{P}} \geq {V}_{\alpha}^{\pi_{k+1}, \mathcal{M}_{k+1}^\beta}-D_{\alpha,\beta}(\mathcal{M}_{k+1}^\beta)
\end{equation}
Since the LHS of (\ref{eq:prop44p1}), i.e., ${V}^{\pi_{k+1}, \mathcal{P}}$ is bigger than the RHS of (\ref{eq:prop44p1}), i.e., ${V}_{\alpha}^{\pi_{k+1}, \mathcal{M}_{k+1}^\beta}-D_{\alpha,\beta}(\mathcal{M}_{k+1}^\beta)$, we can change the inequality into an equation with the RHS plus a update residual $\epsilon_{k+1}$:

\begin{equation}\label{proof:prop44p1}
{V}^{\pi_{k+1}, \mathcal{P}} = {V}_{\alpha}^{\pi_{k+1}, \mathcal{M}_{k+1}^\beta}-D_{\alpha,\beta}(\mathcal{M}_{k+1}^\beta) + \epsilon_{k+1}
\end{equation}
Where

\begin{equation}
\epsilon_{k+1} \triangleq {V}^{\pi_{k+1}, \mathcal{P}} - \left[{V}_{\alpha}^{\pi_{k+1}, \mathcal{M}_{k+1}^\beta} - D_{\alpha,\beta}(\mathcal{M}_{k+1}^\beta)\right]
\end{equation}
With the general pattern equation (\ref{eq:algo-equation}), we have

\begin{equation}\label{proof:prop44p2}
{V}_{\alpha}^{\pi_{k+1}, \mathcal{M}_{k+1}^\beta}-D_{\alpha,\beta}(\mathcal{M}_{k+1}^\beta) \geq {V}^{\pi_k,\mathcal{P}}-D_{\alpha,\beta}(\mathcal{P})
\end{equation}
Since

\begin{align*}
\epsilon_{\mathcal{P}} &= \underset{\phi\in\Phi}{\mathbb{E}}\left[\underset{s,a\sim \pi,\mathcal{P}}{\mathbb{E}}\left[\left\|\mathcal{P}(s, a)-\mathcal{P}(s, a)\right\|\right]\right] \tag{By equation (\ref{def:delta-M})}\\
&=\underset{\phi\in\Phi}{\mathbb{E}}\left[0\right] = 0
\end{align*}
We can show that

\begin{align}
D_{\alpha,\beta}(\mathcal{P}) &= \frac{(1-\beta)\gamma K}{1-\gamma}\epsilon_{\mathcal{P}}+\frac{\alpha(1+\alpha)(1-\beta)}{(1-\alpha)(1-\gamma)}\mathrm{R}_m\notag\\
&=0+\frac{\alpha(1+\alpha)(1-\beta)}{(1-\alpha)(1-\gamma)}\mathrm{R}_m\notag\\
&=\epsilon_\alpha \label{proof:prop44p3}
\end{align}
With equation (\ref{proof:prop44p1}), (\ref{proof:prop44p2}) and (\ref{proof:prop44p3}), it follows that

\begin{equation}
{V}^{\pi_{k+1}, \mathcal{P}}\geq {V}^{\pi_{k}, \mathcal{P}} + \left[\epsilon_{k+1} - \epsilon_\alpha\right]
\end{equation}

Obviously, the update residual $\epsilon_{k+1}$ is much larger than $\epsilon_\alpha$ most of the time during the training period, implying that $\left[\epsilon_{k+1} - \epsilon_\alpha\right]\geq 0$ \textbf{almost surely}, i.e.,

\begin{equation}
{V}^{\pi_{k+1}, \mathcal{P}}\ \overset{\mathrm{a.s.}}{\geq}\ {V}^{\pi_{k}, \mathcal{P}}+0 
\end{equation}
Where ``$\mathrm{a.s.}$'' means ``almost surely''. Then we finally get

\begin{equation}
{V}^{\pi_{0}, \mathcal{P}}\ \overset{\mathrm{a.s.}}{\leq}\ \cdots\ \overset{\mathrm{a.s.}}{\leq}\ {V}^{\pi_{k}, \mathcal{P}}\ \overset{\mathrm{a.s.}}{\leq}\ {V}^{\pi_{k+1}, \mathcal{P}}\ \overset{\mathrm{a.s.}}{\leq}\ \cdots
\end{equation}

\end{proof}

\section{Experiment Details}

\subsection{Environment Settings}

In our experiments, we evaluate our approach on four continuous control benchmark tasks in the Mujoco simulator \cite{todorov2012mujoco}: Hopper, Walker, HalfCheetah and Ant.

\begin{itemize}
    \item \textbf{Hopper}: Make a two-dimensional one-legged robot hop forward as fast as possible.
    \item \textbf{Walker2d}: Make a two-dimensional bipedal robot walk forward as fast as possible.
    \item \textbf{HalfCheetah}: Make a two-dimensional cheetah robot run forward as fast as possible.
    \item \textbf{Ant}: Make a four-legged creature walk forward as fast as possible.
\end{itemize}

\begin{figure}[H]
    \centering
    \subfigure[Hopper]{
        \centering
        \includegraphics[width=0.2\textwidth]{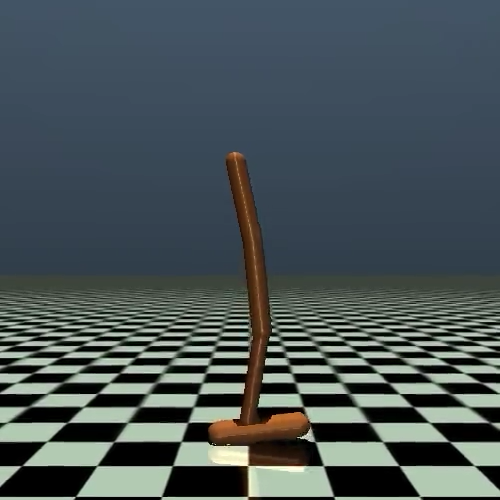}
        \label{fig:hopper}
    }
    \subfigure[Walker2d]{
        \centering
        \includegraphics[width=0.2\textwidth]{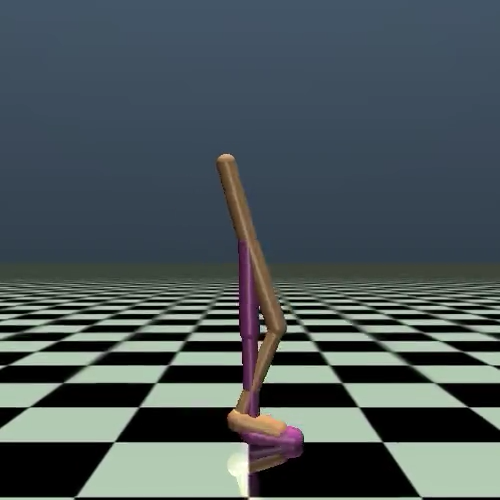}
        \label{fig:walker}
    }
    \subfigure[HalfCheetah]{
        \centering
        \includegraphics[width=0.2\textwidth]{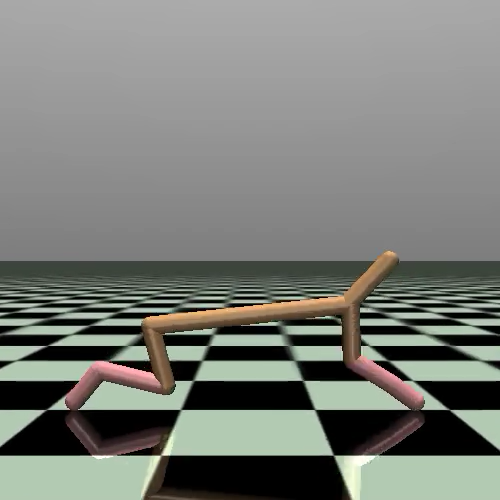}
        \label{fig:cheetah}
    }
    \subfigure[Ant]{
        \centering
        \includegraphics[width=0.2\textwidth]{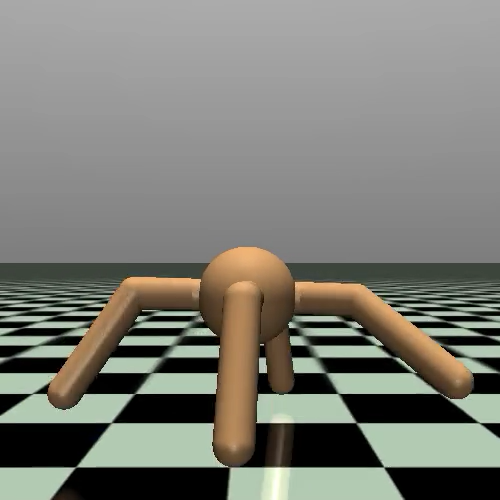}
        \label{fig:ant}
    }
    \caption{Illustrations of the four MuJoCo simulated robot environments used in our experiments.}
    \label{fig:env-figures}
\end{figure}


We also make modifications to the XML model configuration files of MuJoCo environments in our robustness experiments, aiming to evaluate robustness of our MBDP algorithm. More specifically, we scale the standard (OpenAI Gym) friction of each geom part by a coefficient $C_\mathrm{friction}\in [0.5,1.5]$, and scale the standard mass of torso part by a coefficient $C_\mathrm{mass}\in[0.5,1.5]$.

\subsection{Hyperparameter Settings}

\begin{table}[H]
\centering

\begin{adjustbox}{max width=\textwidth}
\begin{tabular}{c|cccccccccc}
\toprule[2pt]
\begin{tabular}[c]{@{}c@{}}environment\\ name\end{tabular} &
  epochs &
  \begin{tabular}[c]{@{}c@{}}env steps\\ per epoch\end{tabular} &
  \begin{tabular}[c]{@{}c@{}}rollout\\ batch\end{tabular} &
  \begin{tabular}[c]{@{}c@{}}policy update\\ per env step\end{tabular} &
  \begin{tabular}[c]{@{}c@{}}model update\\ per env step\end{tabular} &
  $\alpha$ &
  $\beta$ &
  $\gamma$ &
  \begin{tabular}[c]{@{}c@{}}ensemble\\ size\end{tabular} &
  \begin{tabular}[c]{@{}c@{}}network\\ arch\end{tabular} \\
\midrule[1pt]
Hopper      & 120 & 1000 & $10^5$ & 20 & 250 & 0.2 & 0.2 & 0.99 & 10 & \begin{tabular}[c]{@{}c@{}}MLP \\ ($4\times 200$)\end{tabular} \\
Walker2d    & 300 & 1000 & $10^5$ & 20 & 250 & 0.2 & 0.2 & 0.99 & 10 & \begin{tabular}[c]{@{}c@{}}MLP \\ ($4\times 200$)\end{tabular} \\
HalfCheetah & 400 & 1000 & $10^5$ & 40 & 250 & 0.2 & 0.2 & 0.99 & 10 & \begin{tabular}[c]{@{}c@{}}MLP \\ ($4\times 200$)\end{tabular} \\
Ant         & 300 & 1000 & $10^5$ & 20 & 250 & 0.2 & 0.2 & 0.99 & 10 & \begin{tabular}[c]{@{}c@{}}MLP \\ ($4\times 200$)
\end{tabular} \\
\bottomrule[2pt]
\end{tabular}
\end{adjustbox}
\caption{Hyperparameter settings for MBDP results shown in Figure \ref{fig:performance} of the main manuscript.}
\label{tab:hyperparameters}

\end{table}

\subsection{Computational Details}

\begin{table}[H]
\centering
\begin{tabular}{c|c|c}
\toprule[2pt]
CPU                      & GPU                         & RAM   \\
\midrule[1pt]
Intel E5-2680@2.4GHz (56 Cores) & Tesla P40 (24GB) $\times$ 8 & 256GB\\
\bottomrule[2pt]
\end{tabular}
\caption{Computational resources for our experiments.}
\label{tab:computing-resources}
\end{table}

\begin{table}[H]
\centering
\begin{tabular}{c|c|c|c|c}
\toprule[2pt]
Environment Name  & Hopper      & Walker2d    & HalfCheetah & Ant         \\
\midrule[1pt]
Time & $\approx 10$ hours & $\approx 20$ hours & $\approx 32$ hours & $\approx 48$ hours \\
\bottomrule[2pt]
\end{tabular}
\caption{Computing time of each single experiment in different environments.}
\label{tab:run-time}
\end{table}

\section{Empirical Demonstration of Proposition \ref{prop:performance}}

\begin{figure}[H]
    \centering
    \includegraphics[width=\textwidth]{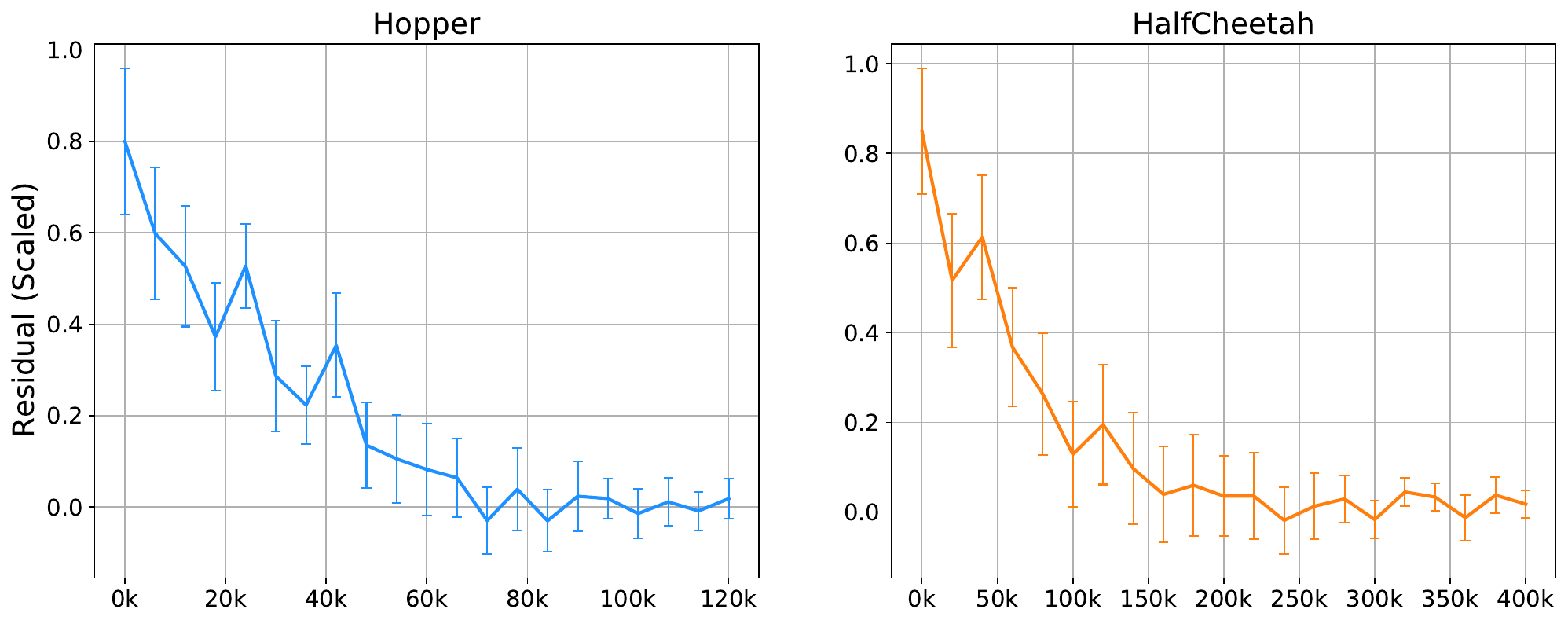}
    \caption{Scaled residual curve in Hopper and HalfCheetah environment.}
    \label{fig:residual}
\end{figure}

The observation results of the residual value is shown in Figure \ref{fig:residual}. The horizontal axis is the number of training epochs, and the vertical axis represents the estimated value (scaled to 1) of $(\epsilon_k-\epsilon_\alpha)$. It can be seen from the figure that $(\epsilon_k-\epsilon_\alpha)$ is greater than 0 most of the time, and occasionally less than 0 when it is close to convergence. This verifies our claim in Section \ref{sec:gua-effi}.


\end{document}